\documentclass[runningheads]{llncs}
\usepackage[T1]{fontenc}
\usepackage{adjustbox}
\usepackage{floatrow}
\usepackage{graphics}

\usepackage[font=small,labelfont=bf]{caption}

\usepackage{graphicx}
\usepackage{xcolor}
\usepackage{array}
\usepackage{amsmath}
\usepackage{amsfonts}
\usepackage{soul}
\begin{document}
\title{Detection of diabetic
retinopathy using longitudinal self-supervised learning}
%
%
\author{Rachid Zeghlache\inst{1,2}
\and
Pierre-Henri Conze\inst{1,3} 
\and
Mostafa El Habib Daho \inst{1,2}
\and
Ramin Tadayoni\inst{5} 
\and
Pascal Massin\inst{5} 
\and
Béatrice Cochener\inst{1,2,4} 
\and 
Gwenolé Quellec\inst{1} 
\and
Mathieu Lamard\inst{1,2} 
}

\authorrunning{R. Zeghlache et al.}
%
\institute{
LaTIM UMR 1101, Inserm, Brest, France \and
University of Western Brittany, Brest, France \and
IMT Atlantique, Brest, France
\and
Ophtalmology Department, CHRU Brest, Brest, France \and
Lariboisière Hospital, AP-HP, Paris, France
}
\maketitle              
\begin{abstract}


Longitudinal imaging is able to capture both static anatomical structures and dynamic changes in disease progression towards earlier and better patient-specific pathology management. However, conventional approaches for detecting diabetic retinopathy (DR) rarely take advantage of longitudinal information to improve DR analysis. In this work, we investigate the benefit of exploiting self-supervised learning with a longitudinal nature for DR diagnosis purposes. We compare different longitudinal self-supervised learning (LSSL) methods to model the disease progression from longitudinal retinal color fundus photographs (CFP) to detect early DR severity changes using a pair of consecutive exams. The experiments were conducted on a longitudinal DR screening dataset with or without those trained encoders (LSSL) acting as a longitudinal pretext task. Results achieve an AUC of 0.875 for the baseline (model trained from scratch) and an AUC of 0.96 (95\% CI: 0.9593-0.9655 DeLong test) with a p-value < 2.2e-16 on early fusion using a simple ResNet alike architecture with frozen LSSL weights, suggesting that the LSSL latent space enables to encode the dynamic of DR progression.


\end{abstract}

\section{Introduction}

\begin{figure}[h!]
\includegraphics[width=\textwidth]{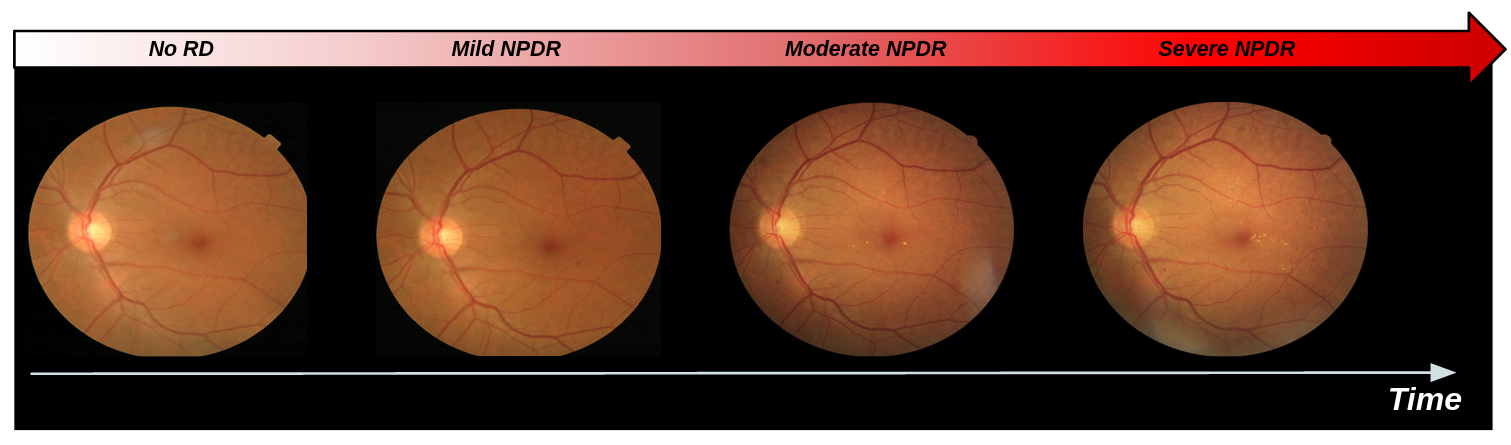}
\caption{Evolution from no DR to severe NPDR for a patient in OPHDIAT \cite{ophdiat} dataset.
} 
\label{fig:progresion_DR}
\end{figure}

According to the International Diabetes Federation, the number of people affected by diabetes is expected to reach 700 million by 2045 \cite{Saeedi2019}. Diabetic retinopathy (DR) affects over one-third of this population and is the leading cause of vision loss worldwide \cite{ogurtsova2017idf}. This happens when the retinal blood vessels are damaged by high blood sugar levels, causing swelling and leakage. In fundus retina images, lesions appear as leaking blood and fluids. Red and bright lesions are the type of lesions  that can be commonly identified during DR screening. The blindness incidence can be reduced if the DR is detected at an early stage. In clinical routine, color fundus photographs (CFP) are employed to identify the morphological changes of the retina by examining the presence of retinal lesions such as microaneurysms, hemorrhages, and soft or hard exudates. The international clinical DR severity scale includes no apparent DR, mild non-proliferative diabetic retinopathy (NPDR), moderate NPDR, severe NPDR, and proliferative diabetic retinopathy (PDR), labeled as grades 0, 1, 2, 3, (illustrated in Fig.\ref{fig:progresion_DR}) and 4. NPDR (grades 1, 2, 3) corresponds to the early-to-middle stage of DR and deals with a progressive microvascular disease characterized by small vessel damages and occlusions. PDR (grade 4) corresponds to the period of potential visual loss which is often due to a massive hemorrhage. Early identification and adequate treatment, particularly in the mild to moderate stage of NPDR, may slow the progression of DR, consequently preventing the establishment of diabetes-related visual impairments and blindness.

In the past years, deep learning has achieved great success in medical image analysis. Many supervised learning techniques based on convolutional neural networks have been proposed to tackle the automated DR grading task \cite{PRATT2016200,QUELLEC2017178,gayathri2020lightweight}. Nevertheless, these approaches rarely take advantage
of longitudinal information. In this direction, Yan et al. \cite{Yan} proposed to exploit a Siamese network with different pre-training and fusion schemes to detect the early stage of RD using longitudinal pairs of CFP acquired from the same patient. Further, self-supervised learning (SSL) held great promise as it can learn robust high-level representations by training on pretext tasks \cite{e24040551} before solving a supervised downstream task. Current self-supervised models are largely based on contrastive learning \cite{liu2020self,https://doi.org/10.48550/arxiv.2002.05709}. However, the choice of the pretext task to learn a good representation is not straightforward, and the application of contrastive learning to medical images is relatively limited. To tackle this, a self-supervised framework using lesion-based contrastive learning was employed for automated diabetic retinopathy (DR) grading \cite{Lesion-based}.

More recently, a new pretext task has appeared for classification purposes in a longitudinal context. Rivail et al. \cite{Rivail2019} proposed a longitudinal self-supervised learning Siamese model trained to predict the time interval between two consecutive longitudinal retinal optical coherence tomography (OCT) acquisitions and thus capturing the disease progression. Zhao et al. \cite{Zhao2021} proposed an auto-encoder named LSSL that takes two consecutive longitudinal scans as inputs. They added to the classic reconstruction term a time alignment term that forces the topology of the latent space to change in the direction of longitudinal changes. An extension of such principle was provided in \cite{Ouyang}. To reach a smooth trajectory field, a dynamic graph in each training batch was computed to define a neighborhood in the latent space for each subject. The graph then connects nearby subjects and enforces their progression directions to be maximally aligned.

In this regard, we aim to use LSSL approaches to capture the disease progression to predict the change between no DR/mild NPDR (grade 0 or 1) and more severe DR (grade $\geq$2) through two consecutive follow-ups. To this end, we explore three methods incorporating current and prior examinations. Finally, a comprehensive evaluation is conducted by comparing these pipelines on the OPHDIAT dataset \cite{ophdiat}. To the best of our knowledge, this work is the first to automatically assess the early DR severity changes between consecutive images using self-supervised learning applied in a longitudinal context.

\section{Methods}

\begin{figure}[h!]
\includegraphics[width=\textwidth]{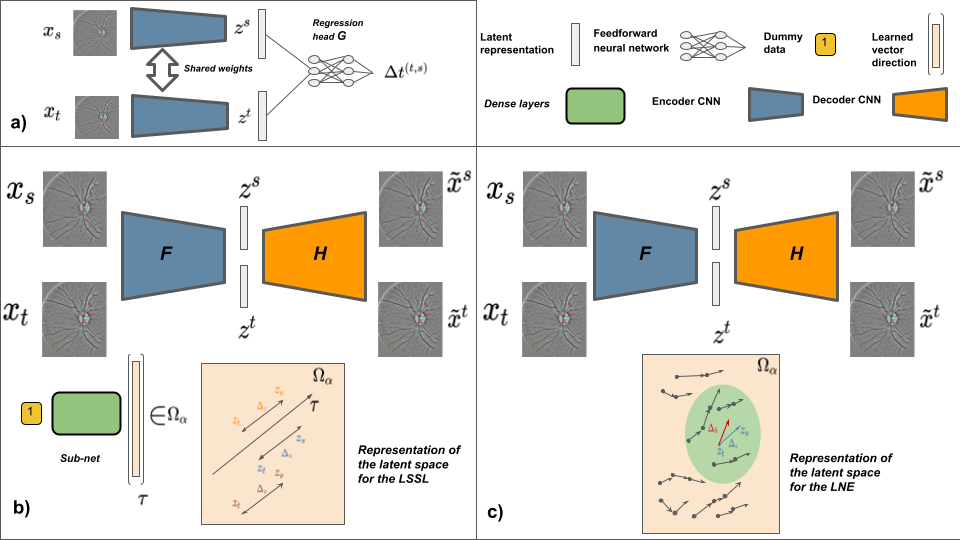}
\caption{The figure a) illustrates to longitudinal Siamese and takes as inputs a pair of consecutive images and predict the time between the examinations. The figure b) represents the longitudinal self-supervised learning which is composed of two independent modules, an AE and dense layers. The AE takes as input the pair of consecutive images and reconstruct the image pairs while the dense layer maps a dummy vector to the direction vector $\tau$. The figure c) corresponds to the LNE, and takes as input the consecutive pairs and build a dynamic graph to align in a neighborhood the subject-specific trajectory vector ($\Delta z$) and the pooled trajectory vector ($\Delta h$) that represents the local progression direction in latent space (green circle).}
\label{fig:overview}
\end{figure}

In this work, we study the use of different longitudinal pretext tasks. We use the encoders trained with those pretext tasks as feature extractors embedded with longitudinal information. The aim is to predict the severity grade change from normal/mild NPDR to more severe DR between a pair of follow-up CFP images. Let $\mathcal{X}$ be the set of subject-specific image pairs for the collection of all CFP images. $\mathcal{X}$ contains all $(x^t, x^s)$ that are from the same subject with $x^t$ scanned before $x^s$. These image pairs are then provided as inputs to an auto-encoder (AE) structure (Fig.\ref{fig:overview}\textit{c}). The latent representations generated by the encoder are denoted by $z^t=F(x^t)$ and $ z^s=F(x^s)$ where $F$ is the encoder. From this encoder, we can define the $\Delta z = (z^s - z^t)$ trajectory vector and then formulate $\Delta z^{(t,s)} = (z^s - z^t) / \Delta t^{(t,s)}$ as the normalized trajectory vector where $\Delta t^{(t,s)}$ is the time interval between the two acquisitions. The decoder $H$ uses the latent representation to reconstruct the input images such that $\tilde{x}^t=H(z^t)$ and $\tilde{x}^s=H(z^s)$. $\mathbf{E}$ denotes the expected value. In what follows, three longitudinal self-supervised learning schemes are further described.

\subsection{Longitudinal Siamese} 

The Siamese network takes the two image pairs $(x^t, x^s)$. These images are encoded into a compact representation ($z^t$, $z^s$) by the encoder network, $F$. A feed forward neural network (denoted $G$) then predicts $\Delta t^{(t,s)}$, the time interval between the pair of CFP images (Fig.\ref{fig:overview}\textit{a}). The regression model is trained by minimizing the following L2 loss: $\parallel G(z^t,z^s) - \Delta t^{(t,s)} \parallel_2^2$.

\subsection{Longitudinal self-supervised learning} 

The longitudinal self-supervised learning (LSSL) exploits a standard AE. The AE is trained with a loss that forces the trajectory vector $\Delta z$ to be aligned with a direction that could rely in the latent space of the AE called $\tau$. This direction is learned  through a sub-network composed of single dense layers which map a dummy data into a vector $\tau \in \Omega_{\alpha}$, the dimension of the latent space. The high-level representation of the network is illustrated in Fig.\ref{fig:overview}\textit{b}. Enforcing the AE to respect this constraint is equivalent to encouraging $\cos\left(\Delta z,\boldsymbol{\tau}\right)$ to be close to 1, i.e. a zero-angle between $\boldsymbol{\tau}$ and the direction of progression in the representation space.

\noindent\textbf{Objective Function.}
\begin{equation}
\mathbf{E}_{(x^t, x^s) \sim \mathcal{X}} \left(\lambda_{rec}\cdot(\parallel x^t - \tilde{x}^t \parallel_2^2 + \parallel x^s - \tilde{x}^s \parallel_2^2)-\lambda_{dir} \cdot \cos(\Delta z,\tau)\right)
\label{eq:loss_1}
\end{equation}

\subsection{Longitudinal neighbourhood embedding}


Longitudinal neighborhood embedding (LNE) is based on the LSSL framework. The main difference is that a directed graph $\mathcal{G}$ is built in each training batch. A pair of sample ($x_{t},x_{s}$) serves as a node in the graph with node representation $\Delta z$. For each node $i$, Euclidean distances to other nodes $j\neq i$ are computed by $D_{i,j} = \parallel z^t_i - z^t_j\parallel_2$. The neighbour size ($N_{nb}$) is the closest nodes of node $i$ form its 1-hop neighbourhood $\mathcal{N}_i$ with edges connected to $i$. The adjacency matrix $A$ for the directed graph ($\mathcal{G}$) is then defined as:

\begin{align*} 
    A_{i,j} := 
   \begin{cases}
    exp(-\frac{D_{i,j}^2}{2\sigma_i^2})  & j \in \mathcal{N}_i\\
    0, & j \notin \mathcal{N}_i
    \end{cases}~. \\
    \mbox{with } \sigma_i := max(D_{i,j \in \mathcal{N}_i}) - min(D_{i,j \in \mathcal{N}_i})
    \label{eqn:adj}
\end{align*}
This matrix regularizes each node's representation by a longitudinal neighbourhood embedding $\Delta h$ pooled from the neighbours' representations. The neighborhood embedding for a node $i$ is computed by:
\begin{equation*}
\Delta h_i := \sum _{j \in \mathcal{N}_i} A_{i,j} O^{-1}_{i,j} \Delta z_j,
\end{equation*} where $O$ is the out-degree matrix of graph $\mathcal{G}$, a diagonal matrix that describes the sum of the weights for outgoing edges at each node. They define $\theta_{\langle \Delta z,\Delta h \rangle}$ the angle between $\Delta z$ and $\Delta h$, and only incite $\cos(\theta_{\langle \Delta z,\Delta h \rangle}) = 1$, i.e., a zero-angle between the subject-specific trajectory vector and the pooled trajectory vector that represents the local progression direction in the latent space (Fig. \ref{fig:overview}\textit{c}). 

\noindent\textbf{Objective Function.} 
\begin{equation}
    \mathbf{E}_{(x^t, x^s) \sim \mathcal{X}} \left(\lambda_{rec} \cdot(\parallel x^t - \tilde{x}^t \parallel_2^2 + \parallel x^s - \tilde{x}^s \parallel_2^2)- \lambda_{dir} \cdot \cos(\theta_{\langle \Delta z,\Delta h \rangle})\right) \label{eq:loss_2}
\end{equation}

\section{Dataset}

The proposed models were trained and evaluated on OPHDIAT \cite{ophdiat}, a large CFP database collected from the Ophthalmology Diabetes Telemedicine network consisting of examinations acquired from 101,383 patients between 2004 and 2017. Within 763,848 interpreted CFP images 673,017 are assigned with a DR severity grade, the others being non-gradable. Image sizes vary from 1440 $\times$ 960 to 3504 $\times$ 2336 pixels. Each examination has at least two images for each eye. Each subject had 2 to 16 scans with an average of 2.99 scans spanning an average time interval of 2.23 years. The age range of the patients is from 9 to 91.  

\noindent\textbf{Image pair selection.} The majority of patients from the OPHDIAT database have multiple images with different fields of view for both eyes. To facilitate the pairing, we propose to select a single image per eye for each examination: we select the one that best characterizes the DR severity grade, as detailed hereafter. For this purpose, we train a standard classifier using the complete dataset that predicts the DR severity grade (5 grades). During the first epoch, we randomly select one image per eye and per examination for the full dataset. After the end of the first epoch with the learned weights of the model, for each image present in every examination, we select the image that gives the highest classification probability. We repeat this process until the selected images by the model converge to a fixed set of images per examination. From the set of selected images, we construct consecutive pairs for each patient and finally obtain 100,033 pairs of images from 26,483 patients. Only 6,690 (6.7\%) pairs have severity grade changes from grade 0 or 1 to grade $\geq 2$ against 93,343 (93.3\%) pairs with severity  changes that lie between grades 0 and 1. The resulting dataset exhibits the following proportions in gender (Male 52\%, Female 48\%) and diabetes type (type 2 69\%, type 1 31\%). This dataset was further divided into training (60\%), validation (20\%), and test (20\%) based on subjects, i.e., images of a single subject belonged to the same split and in a way that preserves the same proportions of examples in each class as observed in the original dataset.  

\noindent\textbf{Image pre-processing.} Image registration is a fundamental pre-processing step for longitudinal analysis \cite{Saha2019}. Therefore, using an affine transformation, we first conducted a registration step to align $x_{t}$ to $x_{s}$. Images are then adaptively cropped to the width of the field of view (i.e., the eye area in the CPF image) and then resized to 256$\times$256. A Gaussian filter estimates the background in each color channel to attenuate the strong intensity variations among the dataset which is then subtracted from the image. Finally, the field of view is eroded by 5\% to eliminate illumination artifacts around the edges. During the training, random resized crops ([0.96, 1.0] as scale range and [0.95, 1.05] as aspect ratio range) are applied for data augmentation purposes. 


\section{Experiments and Results}

\noindent\textbf{Implementation Details.} As it was conducted in \cite{Zhao2021,Ouyang}, we constructed a standard AE for all the compared methods to focus only on the presented methodology and make a fair comparison between approaches, with the hope that using advanced AE structures could lead to better encoding and generalization. In our basic architecture, we employed a stack of $n$ pre-activated residual blocks, where $n$ determines the depth of that scale for the encoder. In each res-block, the residual feature map was calculated using a series of three 3$\times$3 convolutions, the first of which always halves the number of the feature maps employed at the present scale, such that the residual representations live on a lower-dimensional manifold. Our encoder comprises five levels; the first four levels are composed of two residual blocks, and the latter only one residual block. This provides a latent representation of size $64\times4\times4$. The employed decoder is a reverse structure of the encoder. The different networks were trained for 100 epochs by the AdamW optimizer with a learning rate of $5 \times 10^{-4}$, OneCycleLR as scheduler and a weight decay of $10^{-5}$, using an A6000 GPU with the PyTorch framework. The regularization weights were set to $\lambda_{dir}=1.0$ and $\lambda_{rec}=5.0$. A batch size of 64 was used for all models, and a neighbour size $N_{nb}=5$ and $\Delta z^{(t,s)}$ were used for the LNE, as in the original paper \cite{Ouyang}.

\subsection{Comparison of the approaches on the early change detection}

\begin{figure}
\begin{floatrow}
\ffigbox{%
 \includegraphics[width=6cm]{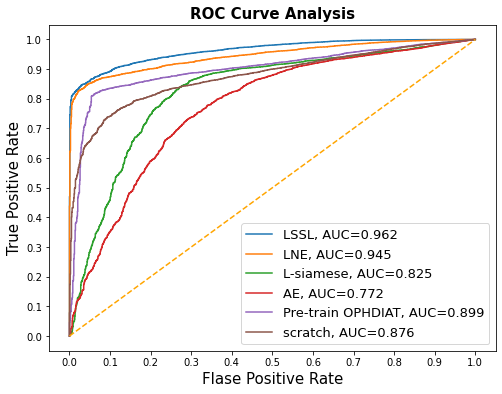}%
}{%
  \caption{ROC Curve Analysis of the compared methods}\label{fig:ROC_curve}%
}
\capbtabbox{%
\begin{adjustbox}{width=\columnwidth,center}%
\begin{tabular}{l c c}
    \cline{2-3} 
      & AUC (95\% CI) & Acc \tabularnewline
    \cline{2-3} 
    Model & & \tabularnewline
    \hline
    No pretrain& 0.8758 (0.8688-0.883) & 0.8379 \tabularnewline
    Pre-train on OPHDIAT& 0.8994 (0.8921-0.9068) &	0.8289 \tabularnewline
    AE& 0.7724 (0.7583-0.7866) & 0.5599 \tabularnewline
    L-siamese \cite{Rivail2019}& 0.8253  (0.8127-0.838) & 0.9354 \tabularnewline
    LSSL \cite{Zhao2021}& \textbf{0.9624} (0.9593-0.9655) & 0.8871 \tabularnewline
    LNE  \cite{Ouyang} & \textbf{0.9448} (0.9412-0.9486)  & 0.8646 \tabularnewline
    \hline
    & & \tabularnewline
    & & \tabularnewline
    & & \tabularnewline
    & & \tabularnewline
    & & \tabularnewline

\end{tabular}
\end{adjustbox}

}{%
\caption{Comparison of the approach on the early change detection with the frozen encoder.}\label{table:table_comparaison_AUC}%
}%
\end{floatrow}
\end{figure}

We evaluate the LSSL encoders on detecting the severity grade change from normal/mild NPDR to more severe DR between a pair of follow-up CFP images. The classifier was constructed as the concatenation of the learned backbone (feature extractor) and a multi-layer perceptron (MLP). The MLP consists of two fully connected layers of dimensions 1024 and 64 with LeakyReLU activation followed by a last single perceptron. Receiving the flattened representation of the trajectory vector $\Delta z$, the MLP predicts a score between 0 and 1.
We compared the area under the receiver operating characteristics curve (AUC) and the accuracy (Acc) in Tab.\ref{table:table_comparaison_AUC} for different pre-training strategies (from scratch, trained on LSSL methods, encoder from a standard AE). We also pre-trained on the OPHDIAT dateset (classification of the DR severity grade) to compare the LSSL pre-training strategies with a conventional pre-training method. The statistical significance was estimated using DeLong’s t-test \cite{Robin2011} to analyze and compare ROC curves. The results in Tab.\ref{table:table_comparaison_AUC} and Fig.\ref{fig:ROC_curve} show the clear superiority of the LSSL encoder, with a statistical significance p-value < 2.2e-16. Due to class imbalance, the Longitudinal-siamese (L-siamese) have a high Acc while exhibiting a lower AUC than the baseline (trained from scratch). 

\newfloatcommand{capbtabbox}{table}[][\FBwidth]

\subsection{Norm of trajectory vector analyze}

\begin{figure}[h!]
\includegraphics[width=\textwidth]{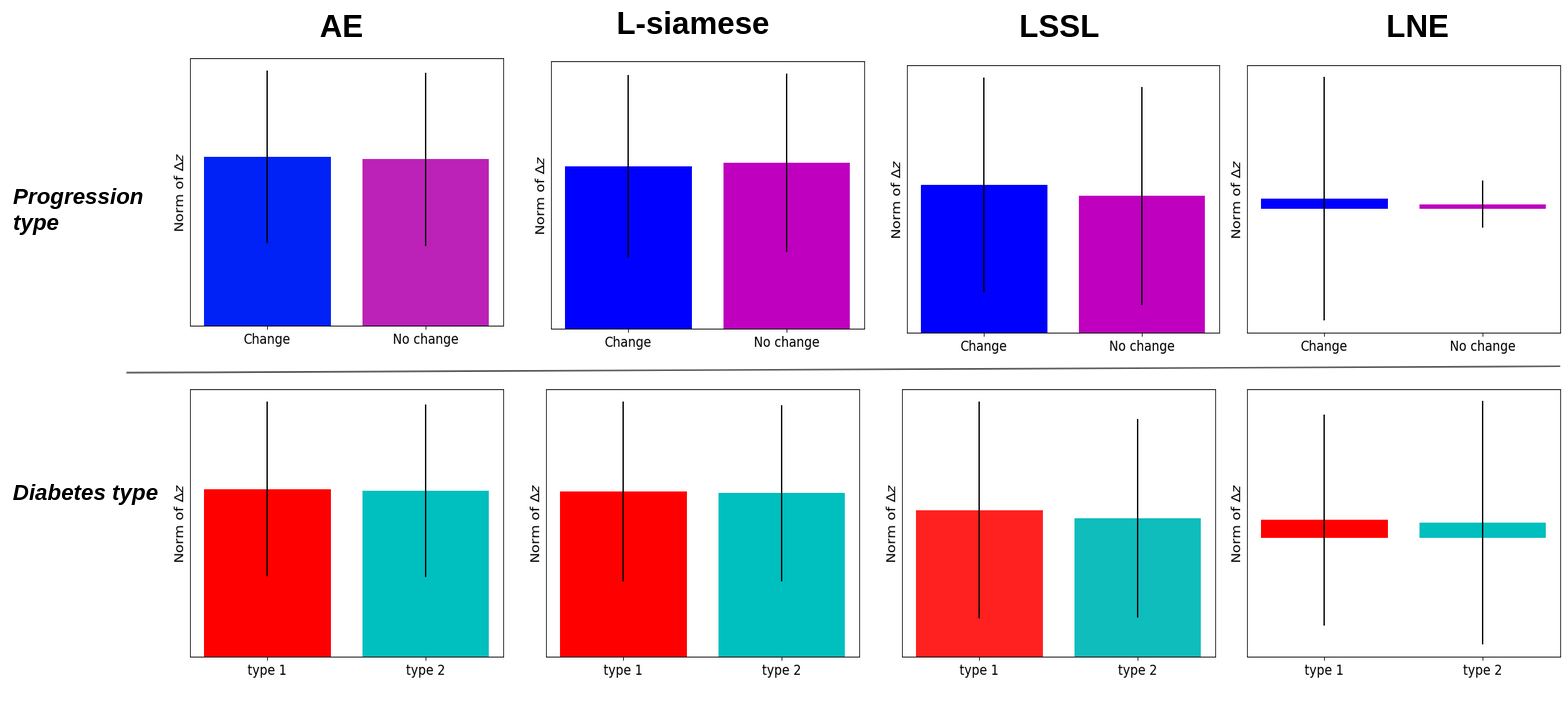}
\caption{Mean of the trajectory vector norm for the different self-supervised method used} 
\label{fig:norm_plot}
\end{figure}

We constructed different histograms in Fig.\ref{fig:norm_plot} representing the mean value of the norm of the trajectory vector with respect to both diabetes type and progression type. According to Fig.\ref{fig:norm_plot}, only the models with the direction alignment loss term are able to capture the change detection in the DR relative to the longitudinal representation. Therefore, we observe in the histogram that the trajectory vector $(\Delta z)$ is able to dissociate the two types of diabetes (t-test p-value < 0.01) and change detection (t-test p-value < 0.01). For the diabetes type, a specific study \cite{Chamard2020} about the OPHDIAT dataset indicates that the DR progression is faster for 
patients with type 1 diabetes. Based on the fact that the $\Delta z$ can be seen as a relative speed, this observation agrees with the histogram plot of the mean of the $\Delta z$ norm represented in Fig. \ref{fig:norm_plot}. We also observed that the norm of the $\Delta z$ vector is lower for the normal stage of the DR than for mild NPDR to more severe. This was expected because only the methods with a direction alignment term in their objective explicitly modeled longitudinal effects, resulting in more informative $\Delta z$. This also implies that simply computing the trajectory vector itself is not enough to force the representation to capture the longitudinal change. 

\section{Discussion}
We applied different LSSL techniques to encode diabetic retinopathy (DR) progression. The accuracy boost, relative to a network trained from scratch or transferred from conventional tasks, demonstrates that longitudinal pre-trained self-supervised representation learns clinically meaningful information. Concerning the limitations of the LSSL methods, we first observe that the models with no time alignment loss perform poorly and provide no evidence of disease progression encoding. Also, we report for the LNE that the normalized trajectory vector for some pairs, that have a large time between examinations, is almost all zeros, which results in a non-informative representation. This could explain the difference between the LSSL and LNE prediction performances. Moreover, during the LSSL and the LNE training, we often faced a plateau with the direction loss alignment. Therefore, we also claim that intensive work should be done regarding the choice of the hyperparameters : constant weights for the losses, latent space size ($\Omega_{\alpha}$), neighbour size ($N_{nb}$). The results concerning quantifying the encoding of the disease progression from the models trained with a time direction alignment are encouraging but not totally clear. As it was mentioned in \cite{Vernhet2021}, one limitation of the LSSL approach pretains to the cosine loss (direction alignment term from equations (\ref{eq:loss_1},\ref{eq:loss_2}) used to encode the longitudinal progression in a specific direction in the latent space and learned while training. The loss only focuses on the correlation with the disease progression timeline but not disentanglement of the disease progression itself. Therefore, a more in-depth analysis of the latent space is required to evaluate if the trajectory vector could be used to find a specific progression trajectory according to patient characteristics (diabetes types, age, DR severity). The pairing and the registration are critical steps in the longitudinal study. As it was previously mentioned, by using a better registration method and exploiting different fusion schemes and backbone architectures, we could get enriched latent representation and, thus, hopefully, better results. Also, the frozen encoders could be transferred to other types of longitudinal problems. In summary, LSSL techniques are quite promising: preliminary results are encouraging, and we expect further improvements. \\

\noindent\textbf{Acknowledgements}
The work takes place in the framework of the ANR RHU project Evired. This work benefits
from State aid managed by the French National Research Agency under “Investissement
d’Avenir” program bearing the reference ANR-18-RHUS-0008







%
%
%
\bibliographystyle{splncs04}
\bibliography{bibliography}

\begin{thebibliography}{10}
\providecommand{\url}[1]{\texttt{#1}}
\providecommand{\urlprefix}{URL }
\providecommand{\doi}[1]{https://doi.org/#1}

\bibitem{e24040551}
Albelwi, S.: Survey on self-supervised learning: Auxiliary pretext tasks and
  contrastive learning methods in imaging. Entropy  \textbf{24}(4) (2022).
  \doi{10.3390/e24040551}, \url{https://www.mdpi.com/1099-4300/24/4/551}

\bibitem{Chamard2020}
Chamard, C., Daien, V., Erginay, A., Gautier, J.F., Villain, M., Tadayoni, R.,
  Carriere, I., Massin, P.: Ten-year incidence and assessment of safe screening
  intervals for diabetic retinopathy: the {OPHDIAT} study. British Journal of
  Ophthalmology  \textbf{105}(3),  432--439 (Jun 2020).
  \doi{10.1136/bjophthalmol-2020-316030},
  \url{https://doi.org/10.1136/bjophthalmol-2020-316030}

\bibitem{https://doi.org/10.48550/arxiv.2002.05709}
Chen, T., Kornblith, S., Norouzi, M., Hinton, G.: A simple framework for
  contrastive learning of visual representations (2020).
  \doi{10.48550/ARXIV.2002.05709}, \url{https://arxiv.org/abs/2002.05709}

\bibitem{gayathri2020lightweight}
Gayathri, S., Gopi, V.P., Palanisamy, P.: A lightweight cnn for diabetic
  retinopathy classification from fundus images. Biomedical Signal Processing
  and Control  \textbf{62},  102115 (2020)

\bibitem{Lesion-based}
Huang, Y., Lin, L., Cheng, P., Lyu, J., Tang, X.: Lesion-based contrastive
  learning for diabetic retinopathy grading from fundus images  (2021).
  \doi{10.48550/ARXIV.2107.08274}, \url{https://arxiv.org/abs/2107.08274}

\bibitem{liu2020self}
Liu, X., Zhang, F., Hou, Z., Wang, Z., Mian, L., Zhang, J., Tang, J.:
  Self-supervised learning: Generative or contrastive. arXiv preprint
  arXiv:2006.08218  \textbf{1}(2) (2020)

\bibitem{ophdiat}
Massin, P., Chabouis, A., Erginay, A., Viens-Bitker, C., Lecleire-Collet, A.,
  Meas, T., Guillausseau, P., Choupot, G., André, B., Denormandie, P.:
  Ophdiat©: A telemedical network screening system for diabetic retinopathy in
  the ile-de-france. Diabetes \& metabolism  \textbf{34},  227--34 (07 2008).
  \doi{10.1016/j.diabet.2007.12.006}

\bibitem{ogurtsova2017idf}
Ogurtsova, K., da~Rocha~Fernandes, J., Huang, Y., Linnenkamp, U., Guariguata,
  L., Cho, N.H., Cavan, D., Shaw, J., Makaroff, L.: {IDF} diabetes atlas:
  Global estimates for the prevalence of diabetes for 2015 and 2040. Diabetes
  Research and Clinical Practice  \textbf{128},  40--50 (2017)

\bibitem{Ouyang}
Ouyang, J., Zhao, Q., Adeli, E., Sullivan, E.V., Pfefferbaum, A., Zaharchuk,
  G., Pohl, K.M.: Self-supervised longitudinal neighbourhood embedding

\bibitem{PRATT2016200}
Pratt, H., Coenen, F., Broadbent, D.M., Harding, S.P., Zheng, Y.: Convolutional
  neural networks for diabetic retinopathy. Procedia Computer Science
  \textbf{90},  200--205 (2016).
  \doi{https://doi.org/10.1016/j.procs.2016.07.014},
  \url{https://www.sciencedirect.com/science/article/pii/S1877050916311929},
  20th Conference on Medical Image Understanding and Analysis (MIUA 2016)

\bibitem{QUELLEC2017178}
Quellec, G., Charrière, K., Boudi, Y., Cochener, B., Lamard, M.: Deep image
  mining for diabetic retinopathy screening. Medical Image Analysis
  \textbf{39},  178--193 (2017).
  \doi{https://doi.org/10.1016/j.media.2017.04.012},
  \url{https://www.sciencedirect.com/science/article/pii/S136184151730066X}

\bibitem{Rivail2019}
Rivail, A., Schmidt-Erfurth, U., Vogel, W.D., Waldstein, S.M., Riedl, S.,
  Grechenig, C., Wu, Z., Bogunovic, H.: Modeling disease progression in retinal
  octs with longitudinal self-supervised learning. Lecture Notes in Computer
  Science (including subseries Lecture Notes in Artificial Intelligence and
  Lecture Notes in Bioinformatics)  \textbf{11843 LNCS},  44--52 (2019).
  \doi{10.1007/978-3-030-32281-6_5}

\bibitem{Robin2011}
Robin, X., Turck, N., Hainard, A., Tiberti, N., Lisacek, F., Sanchez, J.C.,
  M\"{u}ller, M.: {pROC}: an open-source package for r and s+ to analyze and
  compare {ROC} curves. {BMC} Bioinformatics  \textbf{12}(1) (Mar 2011).
  \doi{10.1186/1471-2105-12-77}, \url{https://doi.org/10.1186/1471-2105-12-77}

\bibitem{Saeedi2019}
Saeedi, P., Petersohn, I., Salpea, P., Malanda, B., Karuranga, S., Unwin, N.,
  Colagiuri, S., Guariguata, L., Motala, A.A., Ogurtsova, K., Shaw, J.E.,
  Bright, D., Williams, R.: Global and regional diabetes prevalence estimates
  for 2019 and projections for 2030 and 2045: Results from the international
  diabetes federation diabetes atlas, 9th edition. Diabetes Research and
  Clinical Practice  \textbf{157},  107843 (Nov 2019).
  \doi{10.1016/j.diabres.2019.107843},
  \url{https://doi.org/10.1016/j.diabres.2019.107843}

\bibitem{Saha2019}
Saha, S.K., Xiao, D., Bhuiyan, A., Wong, T.Y., Kanagasingam, Y.: Color fundus
  image registration techniques and applications for automated analysis of
  diabetic retinopathy progression: A review. Biomedical Signal Processing and
  Control  \textbf{47},  288--302 (2019). \doi{10.1016/j.bspc.2018.08.034},
  \url{https://doi.org/10.1016/j.bspc.2018.08.034}

\bibitem{Vernhet2021}
Vernhet, P., Durrleman, S.: Longitudinal self-supervision to disentangle
  inter-patient variability pp. 231--241 (2021).
  \doi{10.1007/978-3-030-87196-3}

\bibitem{Yan}
Yan, Y., henri Conze, P.: Longitudinal detection of diabetic retinopathy early
  severity grade  \textbf{3},  11--20. \doi{10.1007/978-3-030-87000-3}

\bibitem{Zhao2021}
Zhao, Q., Liu, Z., Adeli, E., Pohl, K.M.: Longitudinal self-supervised
  learning. Medical Image Analysis  \textbf{71} (2021).
  \doi{10.1016/j.media.2021.102051}

\end{thebibliography}
%





\end{document}